\documentclass[letterpaper]{article} 
\usepackage{aaai2026}  
\usepackage{times}  
\usepackage{helvet}  
\usepackage{courier}  
\usepackage[hyphens]{url}  
\usepackage{graphicx} 
\usepackage{natbib}  
\usepackage{caption}
\usepackage{algorithm}
\usepackage{algorithmic}
\usepackage{amsmath}
\usepackage{makecell}
\usepackage{multirow}
\usepackage{subcaption}
\usepackage{booktabs}
\usepackage[table]{xcolor}
\usepackage{enumerate}

\usepackage{circledsteps}

\usepackage{tikz}

\newcommand{\calMcirc}{%
  \tikz[baseline=(M.base)]{
    \node[draw,circle,inner sep=0.1pt, line width=1pt](M){\scriptsize$\mathbf{M}$};
  }%
}

\newcommand{\calRcirc}{%
  \tikz[baseline=(M.base)]{
    \node[draw,circle,inner sep=0.1pt, line width=1pt](R){\scriptsize$\mathbf{R}$};
  }%
}

\newcommand{\calKcirc}{%
  \tikz[baseline=(M.base)]{
    \node[draw,circle,inner sep=0.1pt, line width=1pt](K){\scriptsize$\mathbf{K}$};
  }%
}

\usepackage{newfloat}
\usepackage{listings}
\DeclareCaptionStyle{ruled}{labelfont=normalfont,labelsep=colon,strut=off}
\floatstyle{ruled}
\newfloat{listing}{tb}{lst}{}
\floatname{listing}{Listing}

\title{Towards Benchmarking Privacy Vulnerabilities in Selective Forgetting \\ with Large Language Models}

\author{
    Wei Qian\equalcontrib\textsuperscript{\rm 1},
    Chenxu Zhao\equalcontrib\textsuperscript{\rm 1},
    Yangyi Li\textsuperscript{\rm 1},
    Mengdi Huai\textsuperscript{\rm 1}
}
\affiliations{
    \textsuperscript{\rm 1}Iowa State University\\
    \{wqi, cxzhao, liyangyi, mdhuai\}@iastate.edu
}

\begin{document}

\maketitle

\begin{abstract}
The rapid advancements in artificial intelligence (AI) have primarily focused on the process of learning from data to acquire knowledgeable learning systems. As these systems are increasingly deployed in critical areas, ensuring their privacy and alignment with human values is paramount. Recently, \emph{selective forgetting} (also known as \emph{machine unlearning}) has shown promise for privacy and data removal tasks, and has emerged as a transformative paradigm shift in the field of AI. It refers to the ability of a model to selectively erase the influence of previously seen data, which is especially important for compliance with modern data protection regulations and for aligning models with human values. Despite its promise, selective forgetting raises significant privacy concerns, especially when the data involved come from sensitive domains. While new unlearning-induced privacy attacks are continuously proposed, each is shown to outperform its predecessors using different experimental settings, which can lead to overly optimistic and potentially unfair assessments that may disproportionately favor one particular attack over the others. In this work, we present the first comprehensive benchmark for evaluating privacy vulnerabilities in selective forgetting. We extensively investigate privacy vulnerabilities of machine unlearning techniques and benchmark privacy leakage across a wide range of victim data, state-of-the-art unlearning privacy attacks, unlearning methods, and model architectures. We systematically evaluate and identify critical factors related to unlearning-induced privacy leakage. With our novel insights, we aim to provide a standardized tool for practitioners seeking to deploy customized unlearning applications with faithful privacy assessments.
\end{abstract}

\section{Introduction}
\label{sec:intro}

In recent years, artificial intelligence (AI) has revolutionized nearly every aspect of modern life. In an era of AI, the primary challenge is enabling models to acquire broad knowledge effectively. However, the training datasets employed in training these models often contain sensitive information encompassing private and copyrighted content \cite{bu2024pre,mueller2024llms,Chu_Song_Yang_2024, wei2024evaluating}. This situation raises significant risks of sensitive data leakage, directly conflicting with the growing legislative emphasis on the ``right to be forgotten'' \cite{ccpa2019, regulation2018general}. Instances such as the proliferation of copyright infringement cases post the release of models \cite{rombach2022high}, and The New York Times’s lawsuit against OpenAI for content leakage \cite{hadero2023new}, underscore the urgency of addressing these issues.

In response to these challenges, \emph{selective forgetting} (also referred to as \emph{machine unlearning})~\cite{li2025machine,zhang2024negative, qian2023towards,zhao2023static,qian2022patient,bourtoule2021machine} has emerged as a promising solution. Selective forgetting aims to compel models to forget sensitive information without retraining, thereby eliminating the risk of content leakage. In contrast, retraining models from scratch to accommodate deletions is impractical due to the extensive computational resources required. Machine unlearning aims to remove the influence of \emph{the requested unlearning data} from a pre-trained model, producing an unlearned model that approximates one retrained from scratch using only \emph{the retain data} (i.e., the original training data excluding the unlearning set). Recent work also leverages conformal prediction~\cite{li2025quantifying} to quantify forgetting uncertainty, leading to more rigorous unlearning~\cite{alkhatib2025conformal}. Machine unlearning not only aids in meeting regulatory requirements but also enhances learning systems’ privacy protection by sensitive data attacks.

However, adopting machine unlearning techniques may not always provide the anticipated privacy protections, and could even introduce new privacy vulnerabilities. First, for the requested unlearning data, machine unlearning naturally generates two versions of machine learning models, namely the original model and the unlearned model, which differ due to the deletion of the unlearning data. This discrepancy can inadvertently leak information about the unlearned data. Additionally, the unlearned model alone may still retain residual privacy risks related to the requested unlearning data due to incomplete forgetting. Moreover, the privacy of the unlearning data may be further compromised when the unlearned model is subjected to future deployment scenarios such as fine-tuning, which may reactivate or amplify memorized knowledge. Even worse, beyond the privacy risks of the unlearning data, selective forgetting may also influence the retain data, potentially altering their privacy exposure through model shifts or malicious fingerprints. These concerns highlight that selective forgetting may introduce new privacy attack surfaces that adversaries can exploit, potentially undermining the guarantees associated with unlearning requests or compromising the privacy of other data.

Currently, many works have been proposed to investigate privacy risks stemming from selective forgetting~\cite{hu2025jogging,lucki2025adversarial,zhang2025catastrophic,yuan2025towards,wang2025tape,hu2024learn, carlini2022privacy,lu2022label,chen2021machine}. For example, \cite{hu2024learn} examines data reconstruction attacks (DRAs) on unlearning data by leveraging the discrepancy between the pre-trained and unlearned models. However, there still lacks a structured understanding of the empirical privacy risks of machine unlearning techniques. Without a clear understanding of the practical risks, practitioners are left with little guidance on how to safely and privately apply machine unlearning techniques in privacy-sensitive settings. Additionally, existing unlearning-induced privacy attacks are typically evaluated under disparate experimental settings, with varying experimental settings. As a result, each of them is often shown to outperform prior methods under its own tailored conditions, leading to overly optimistic and potentially inconsistent evaluations that may unfairly favor certain attacks. Consequently, an in-depth investigation into the effectiveness of unlearning-induced privacy vulnerabilities in a standard and reproducible experimental setting is missing.

To address these limitations, we in this paper introduce the first comprehensive benchmark \textbf{PrivUB}, i.e., \textbf{Priv}acy Vulnerabilities in Machine \textbf{U}nlearning \textbf{B}enchmark. This work makes four major contributions: 

\noindent (1) We present the first benchmark that systematically evaluates existing privacy vulnerabilities introduced by machine unlearning. Our benchmark emphasizes the importance of aligning privacy guarantees with human intent, highlighting gaps between technical implementations and user expectations. Our benchmark reveals fundamental challenges in unlearning and provides a critical foundation for understanding its implications in the context of emerging data protection regulations and broader challenges in AI alignment.

\noindent (2) We instantiate a structured taxonomy of privacy vulnerabilities in machine unlearning by implementing representative attacks across key dimensions, including privacy vulnerability type, victim data type, victim model type, and attacking tool. Each is grounded with a specific threat model.

\noindent (3) We evaluate existing defense methods targeting privacy risks in machine unlearning, analyzing their effectiveness in a structured manner across different types of privacy vulnerabilities, victim data, victim model, and attacking tool.

\noindent (4) Through extensive empirical studies, we conduct a comprehensive evaluation covering 21 unlearning-induced privacy attack and defense methods in machine unlearning, 11 real-world datasets, 10 mainstream models, 10 popular unlearning techniques, and 10 task-specific evaluation metrics.

We present a thorough analysis of the above evaluations from different perspectives to examine privacy vulnerabilities introduced by selective forgetting. Our key findings include: (1) Combining multiple attacking tools (including perturbing unlearned model and perturbing unlearned data) can improve attack effectiveness. (2) The attacking tools of perturbing unlearned data designed for knowledge leakage attacks can be utilized to further enhance the performance of membership inference attacks (MIAs). (3) The privacy risks caused by the fine-tuning method are more severe than those caused by the model quantization method. (4) Existing privacy attacks, with proper adaptation, can be successfully generalized across model types. Notably, we find that attacks originally developed for deep learning models can be applied to large language models (LLMs), and vice versa, while maintaining strong performance. (5) Existing defenses against privacy vulnerabilities generally lack robustness. In particular, some defenses are highly sensitive to the number of attack samples, leading to inconsistent protection.

\section{Related Work}
\label{sec:background}

The rapid development of machine learning models has significantly benefited various applications. However, their increasing deployment has raised serious privacy concerns, particularly in sensitive domains such as healthcare and finance. Notably, models often unintentionally memorize their training data, going beyond merely learning the general patterns within the data. This behavior makes models vulnerable to various privacy attacks, including membership inference attacks~\cite{ zhao2025membership,carlini2022membership,chen2021machine}, data reconstruction attacks~\cite{hu2024learn,du2024textual, yuan2023pseudo}, and knowledge leakage attacks~\cite{hu2025jogging,lucki2025adversarial,yuan2025towards}.

\begin{table*}[t!]
\small
\centering
\begin{tabular}{c@{\hspace{5pt}}c@{\hspace{5pt}}c@{\hspace{5pt}}c@{\hspace{5pt}}c@{\hspace{5pt}}c@{\hspace{5pt}}c@{\hspace{5pt}}c@{\hspace{5pt}}c} 
\hline
\multirow{3}{*}{\textbf{\makecell{Privacy \\ vulnerability \\ type}}} &
\multirow{3}{*}{\textbf{\makecell{Victim data \\ type}}} &
\multirow{3}{*}{\textbf{\makecell{Victim model \\ type}}} &
\multirow{3}{*}{\textbf{\makecell{Attacking \\ tool}}} &
\multirow{3}{*}{\textbf{Paper}} &
\multicolumn{3}{c}{\textbf{Threat model}} &
\multirow{3}{*}{\textbf{\makecell{Architecture}}} \\
\cline{6-8}
& & & & &
\textbf{\makecell{Model access \\ information}} &
\textbf{\makecell{A}} &
\textbf{\makecell{V}} \\
\hline
\multirow{6}{*}{\makecell{Membership \\ inference}} & \multirow{4}{*}{\makecell{Unlearning \\ data}} & \multirow{4}{*}{\makecell{Pre-trained \\ \makecell{and unlearned \\ models}}} & \multirow{4}{*}{\makecell{Pre-trained and \\ \makecell{unlearned model \\ discrepancy}}} & \cite{chen2021machine} & \makecell{\makecell{Posterior/\\Top-k posterior/}\\Label-only} & Yes & Yes & Deep learning \\
& & & & \cite{lu2022label} & Label only & Yes & Yes & Deep learning \\
& & & & \cite{lu2022fp} & Label only & Yes & Yes & Deep learning \\
& & & & \cite{du2024textual} & Loss & No & No & LLM \\
\cline{2-9}
& \multirow{2}{*}{\makecell{Retain \\ data}} & \multirow{2}{*}{\makecell{Unlearned \\ model}} & \multirow{2}{*}{\makecell{Malicious \\ unlearning subset}}  & \cite{carlini2022privacy} & Model weights & No & Yes & Deep learning \\
& & & & \cite{gu2024auditing} & Model weights & No & Yes & Deep learning \\
\hline
\multirow{3}{*}{\makecell{Data \\ reconstruction}} & \multirow{3}{*}{\makecell{Unlearning \\ data}} & \multirow{3}{*}{\makecell{Pre-trained \\ \makecell{and unlearned \\ models}}} & \multirow{3}{*}{\makecell{Pre-trained and \\ \makecell{unlearned model \\ discrepancy}}} & \cite{hu2024learn} & Model weights & No & Yes & Deep learning \\
& & & & \cite{du2024textual} & Model weights & No & Yes & LLM \\
& & & & \cite{wang2025tape} & Posterior & Yes & Yes & Deep learning \\
\hline
\multirow{8}{*}{\makecell{Knowledge \\ leakage}} & \multirow{8}{*}{\makecell{Unlearning \\ data}} & \multirow{4}{*}{\makecell{Fine-tuned \\ model}} & \multirow{4}{*}{\makecell{Perturbing \\ unlearned model}} & \cite{doshi2024does} & Model weights & No & Yes & LLM \\
& & & & \cite{hu2025jogging} & Model weights & No & Yes & LLM \\
& & & & \cite{zhang2025catastrophic} & Model weights & No & Yes & LLM \\
& & & & \cite{lucki2025adversarial} & Model weights & No & Yes & LLM \\
\cline{3-9}
& & \multirow{4}{*}{\makecell{Unlearned \\ model}} & \multirow{4}{*}{\makecell{Perturbing \\ unlearned data}} & \cite{doshi2024does} & Model weights & No & Yes & LLM \\
& & & & \cite{xuan2025unlearning} & Model weights & No & Yes & Deep learning \\
& & & & \cite{hsu2025are} & Model weights & No & Yes & Deep learning \\
& & & & \cite{yuan2025towards} & Model weights & No & Yes & LLM \\
\hline
\end{tabular}
\caption{Categories of existing privacy vulnerabilities in selective forgetting. A: auxiliary dataset; V: victim model architecture.}
\label{tab:CategoryAttack}
\end{table*}

Currently, many privacy benchmarks have been proposed to investigate privacy risks associated with machine learning models~\cite{niu2025comparing,wen2025sok,chen2025survey, zhu2024privauditor,li2023privlm,song2021systematic}. For example, \cite{niu2025comparing} presents a systematic comparison of various membership inference attacks using carefully designed evaluation scenarios. Rigorous privacy evaluation is essential for identifying vulnerabilities in models, and developing a comprehensive understanding of existing research gaps and potential mitigation strategies, thereby promoting alignment with privacy principles and human values. In this work, we aim to benchmark privacy vulnerabilities in selective forgetting. This is the first benchmark to systematically study the unlearning-induced privacy attacks and defenses.

\section{Benchmark Framework}

\subsection{Setup of Privacy Evaluation}

Let \(f(\cdot; \theta)\) denote the pre-trained model, where \(\theta \in \Theta\) denotes the model parameters. Let \(\mathcal{D} = \{(x_i, y_i)\}_{i=1}^n\) denote a dataset of \(n\) samples drawn from an underlying distribution \(\mathcal{P}\) over \(\mathcal{X} \times \mathcal{Y}\). Note that \emph{during the training phase}, a training algorithm \(T\) maps the training dataset \(\mathcal{D}\) to parameters \(\theta\), yielding the pre-trained model. During \emph{the unlearning phase}, the pre-trained model is updated by an unlearning algorithm \(U\), which aims to remove the influence of the requested unlearning data, which is typically a subset of the training data \(\mathcal{D}\). During \emph{the deployment phase}, the model can be further modified via a fine-tuning procedure \(F\) on task-specific data, or a quantization operator \(Q\), which compresses the parameters \(\theta\) for efficient inference. Below, we elaborate the unlearning phase and the deployment.

Note that the goal of machine unlearning is to remove the influence of a designated subset of the training data from a pre-trained model via the targeted unlearning process. Let \(\mathcal{D}_u \subset \mathcal{D}\) denote the subset of data to be unlearned, with \(|\mathcal{D}_u| = m\). Given a model with parameters \(\theta \in \Theta\), obtained via training on \(\mathcal{D}\), an unlearning algorithm \(U: \Theta \times (\mathcal{X} \times \mathcal{Y})^n \times (\mathcal{X} \times \mathcal{Y})^m \to \Theta\) maps the pre-trained model, the full training dataset \(\mathcal{D}\), and the unlearning data \(\mathcal{D}_u\) to an updated model \(\theta_u \in \Theta\). We denote this unlearning process as \(\theta_u \sim U(\theta, \mathcal{D}, \mathcal{D}_u)\). The unlearning objective is to ensure that the resulting model \(\theta_u\) is indistinguishable from a model retrained from scratch on \emph{the retain data} \(\mathcal{D}_r = \mathcal{D} \setminus \mathcal{D}_u\), i.e., \(\theta_r \sim T(\mathcal{D}_r)\), where \(T\) is the training algorithm. This requirement is formally captured by the following condition: for any measurable subset \(\mathcal{B} \subseteq \Theta\), the distributions of the unlearning and retraining procedures should satisfy: \(P(U(T(\mathcal{D}), \mathcal{D},\mathcal{D}_u) \in \mathcal{B}) \leq e^\epsilon P(T(\mathcal{D}_r)  \in \mathcal{B})+\delta, \text { and } P(T(\mathcal{D}_r) \in \mathcal{B}) \leq e^\epsilon P(U(T(\mathcal{D}), \mathcal{D},\mathcal{D}_u) \in \mathcal{B})+\delta\), where \(\epsilon, \delta > 0\) are tolerance parameters controlling the degree of approximation \cite{guo2019certified}. Based on the strength of this guarantee, unlearning algorithms are typically categorized into: \emph{exact unlearning} \cite{bourtoule2021machine} and \emph{approximate unlearning} \cite{kurmanji2023towards}. Exact unlearning requires that \(U(T(\mathcal{D}), \mathcal{D}, \mathcal{D}_u)\) and \(T(\mathcal{D}_r)\) follow the same distribution, corresponding to the ideal case where \(\epsilon = \delta = 0\). In contrast, approximate unlearning relaxes this strict requirement, and allows bounded statistical divergence controlled by \(\epsilon\) and \(\delta\).

For unlearning-induced privacy vulnerabilities during the deployment phase, there are two procedures: model fine-tuning~\cite{hu2022lora} and model quantization \cite{zhang2025catastrophic}. Fine-tuning aims to enhance task-specific performance. Specifically, given a fine-tuning dataset \(\mathcal{D}_{ft} \subseteq \mathcal{X} \times \mathcal{Y}\) of size \(z\), the unlearned model is further adapted using a fine-tuning algorithm \(F: \Theta \times (\mathcal{X} \times \mathcal{Y})^z \rightarrow \Theta\), which updates the unlearned model \(\theta_u\) based on \(\mathcal{D}_{ft}\). Let \(f(\cdot; \theta_{ft})\) denote the resulting fine-tuned model. Additionally, model quantization is applied to reduce the model's memory footprint and improve inference efficiency \cite{zhang2025catastrophic}. Let \(Q: \Theta \to \Theta\) denote a quantization operator that maps full-precision model parameters to a low-precision representation. Given parameters \(\theta_u \in \Theta\), the quantized model is defined as \(f(\cdot; \theta_q)\), where \(\theta_q \sim Q(\theta_u)\).

\begin{figure}[htbp]
    \centering
    \includegraphics[width=0.45\textwidth]{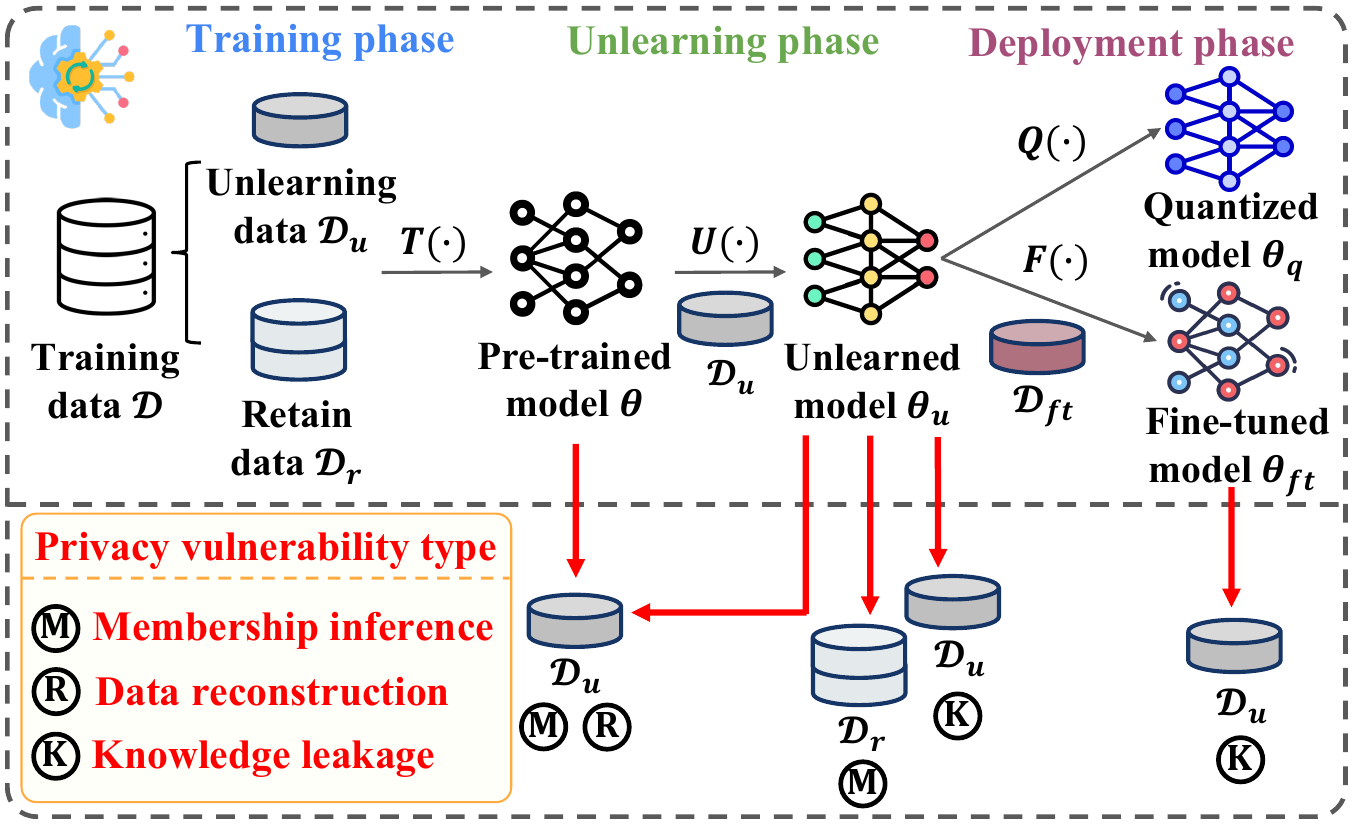} 
    \caption{Privacy vulnerabilities in machine unlearning.}
    \label{fig:PrivacySystem}
\end{figure}

\begin{table*}[t!]
\small
\centering
\begin{tabular}{c@{\hspace{5pt}}c@{\hspace{5pt}}c@{\hspace{5pt}}c@{\hspace{5pt}}c@{\hspace{5pt}}c@{\hspace{5pt}}c} 
\hline
\multirow{2}{*}{\textbf{Paper}} & \multicolumn{4}{c}{\textbf{Defense setting}} & \multirow{2}{*}{\textbf{\makecell{Architecture}}} \\ 
\cline{2-5} & \textbf{Privacy vulnerability type} & \textbf{Victim data type} & \textbf{Victim model type} & \textbf{Attacking tool} \\
\hline
\cite{wang2025crfu} & Membership inference & Unlearning data & \makecell{Pre-trained and \\ unlearned models} & \makecell{Pre-trained and unlearned \\ model discrepancy} & Deep learning \\
\hline
\cite{yuan2025towards} & Knowledge leakage & Unlearning data & Unlearned model & Perturbing unlearned data & LLM \\
\hline
\cite{wang2025crfu} & Data reconstruction & Unlearning data & \makecell{Pre-trained and \\ unlearned models} & \makecell{Pre-trained and unlearned \\ model discrepancy} & Deep learning \\
\hline 
\cite{fan2025towards} & Knowledge leakage & Unlearning data & Fine-tuned model & Perturbing unlearned model & LLM \\
\hline
\cite{tamirisa2025tamperresistant}  & Knowledge leakage & Unlearning data & Fine-tuned model & Perturbing unlearned model & LLM \\
\hline
\end{tabular}
\caption{Categories of existing defenses against privacy vulnerabilities in selective forgetting.}
\label{tab:UnlPrivacyDefense}
\end{table*}

\subsection{Benchmark Design for Unlearning Privacy Risks}
In this section, we detail unlearning-induced privacy vulnerabilities evaluated in our benchmark. As shown in Fig.~\ref{fig:PrivacySystem}, our benchmark evaluates three types of unlearning-induced privacy vulnerabilities: \emph{membership inference attacks} (\calMcirc), \emph{data reconstruction attacks} (\calRcirc), and \emph{knowledge leakage attacks} (\calKcirc). Additionally, our benchmark considers two different victim data types: the unlearning data (\(\mathcal{D}_u\)) and the retain data (\(\mathcal{D}_r\)). Based on this, in Table~\ref{tab:CategoryAttack}, we categorize existing unlearning-induced privacy vulnerabilities along key different dimensions: \emph{privacy vulnerability type}, \emph{victim data type}, \emph{victim model type}, \emph{attacking tool}, \emph{threat model}, and \emph{model architecture}. Below, we summarize the privacy vulnerabilities in selective forgetting.

\textbf{(1) Membership inference attacks for the unlearning data \(\mathcal{D}_u\).} Here, attackers aim to train a membership inference classifier \(M_{1}\) that outputs a binary prediction: 1 if the input was included in the training data \(\mathcal{D}\) of the pre-trained model \(\theta\) and subsequently removed through unlearning, and 0 otherwise \cite{lu2022label,chen2021machine}. To achieve this, attackers aim to characterize the predictive discrepancies between the pre-trained model \(\theta\) and the unlearned model \(\theta_u\) for both members and non-members, leveraging queries under varying levels of model access as the attacking tool. For example, \cite{chen2021machine} assumes access to an auxiliary dataset \(\mathcal{D}_{{aux}}\) and trains shadow models using the same victim model architecture. These shadow models are queried with auxiliary dataset \(\mathcal{D}_{aux}\) to generate full posterior responses, which are then used to train the classifier.

\textbf{(2) Membership inference attacks for the retain data \(\mathcal{D}_r\).} In this category, attackers aim to infer the membership information of samples in the retain dataset \(\mathcal{D}_r=\mathcal{D} \backslash \mathcal{D}_u\) using the unlearned model \(\theta_u\) \cite{gu2024auditing,carlini2022privacy}. For example, \cite{carlini2022privacy} introduces a privacy scoring method to rank the training dataset \(\mathcal{D}\) and select the least private instances as the malicious unlearning subset \(\mathcal{D}_u\) as the attacking tool. The removal of such a subset increases the membership vulnerability of \(\mathcal{D}_r\)  in the resulting model \(\theta_u\), which can be measured using \(M_{2}\).

\textbf{(3) Data reconstruction attacks for the unlearning data \(\mathcal{D}_u\).} The goal of attackers is to train a reconstruction model \(R\) that recovers unlearned data \(\mathcal{D}_u\) from the unlearned model \(\theta_u\), leveraging the model discrepancy between pre-trained model \(\theta\) and unlearned model \(\theta_u\) as the attacking tool~\cite{hu2024learn,du2024textual}. This represents the most severe form of privacy leakage. For example, \cite{hu2024learn} assumes white-box access and proposes matching the gradients of candidate reconstruction inputs to the difference between \(\theta\) and \(\theta_u\), thereby guiding the reconstruction.

\begin{table*}[t!]
\small
\centering
\begin{tabular}{@{\hspace{5pt}}c@{\hspace{5pt}}c|c@{\hspace{5pt}}c|c@{\hspace{5pt}}c|c@{\hspace{5pt}}c}
\hline
\multirow{2}{*}{\textbf{\makecell{Model access \\ information}}} 
& \multirow{2}{*}{\textbf{Method}}
& \multicolumn{2}{c|}{\textbf{Chest X-Ray}} 
& \multicolumn{2}{c|}{\textbf{CelebA}}
& \multicolumn{2}{c}{\textbf{CIFAR-10}}\\
\cline{3-8}
& & \textbf{MIA Acc} \(\uparrow\) & \textbf{AUC} \(\uparrow\) & \textbf{MIA Acc} \(\uparrow\) & \textbf{AUC} \(\uparrow\) & \textbf{MIA Acc} \(\uparrow\) & \textbf{AUC} \(\uparrow\) \\
\hline
\multirow{2}{*}{Posterior} & Basic MIA  & 0.500 $\pm$ 0.020 & 0.482 $\pm$ 0.018 & 0.503 $\pm$ 0.004 & 0.517 $\pm$ 0.029 & 0.502 $\pm$ 0.015 & 0.486 $\pm$ 0.031 \\
& \cite{chen2021machine} & 0.567 $\pm$ 0.013 & 0.600 $\pm$ 0.022 & 0.663 $\pm$ 0.029 & 0.723 $\pm$ 0.033 & 0.607 $\pm$ 0.043 & 0.660 $\pm$ 0.041 \\
\hline
\multirow{2}{*}{Top-k posterior} & Basic MIA & 0.517 $\pm$ 0.008 & 0.494 $\pm$ 0.059 & 0.510 $\pm$ 0.013 & 0.543 $\pm$ 0.028 & 0.482 $\pm$ 0.004 & 0.501 $\pm$ 0.003 \\
& \cite{chen2021machine} & 0.563 $\pm$ 0.012 & 0.557 $\pm$ 0.043 & 0.628 $\pm$ 0.025 & 0.729 $\pm$ 0.018 & 0.605 $\pm$ 0.013 &  0.633 $\pm$ 0.009 \\
\hline
\multirow{4}{*}{Label-only} & Basic MIA & 0.500 $\pm$ 0.000 & 0.502 $\pm$ 0.006 & 0.532 $\pm$ 0.008 & 0.532 $\pm$ 0.008 & 0.505 $\pm$ 0.015 & 0.491 $\pm$ 0.007 \\
& \cite{chen2021machine} & 0.502 $\pm$ 0.015 & 0.504 $\pm$ 0.012 & 0.528 $\pm$ 0.007 & 0.543 $\pm$ 0.013 & 0.502 $\pm$ 0.010 & 0.519 $\pm$ 0.000 \\
& \cite{lu2022label} & 0.813 $\pm$ 0.014 & 0.793 $\pm$ 0.056 & 0.680 $\pm$ 0.009 & 0.725 $\pm$ 0.014 & 0.937 $\pm$ 0.014 & 0.952 $\pm$ 0.001 \\
& \cite{lu2022fp} & 0.805 $\pm$ 0.020 & 0.772 $\pm$ 0.027 & 0.632 $\pm$ 0.048 & 0.699 $\pm$ 0.065 & 0.910 $\pm$ 0.013 & 0.933 $\pm$ 0.016 \\
\hline
\end{tabular}
\caption{Comparisons of membership inference attacks for unlearning data.}
\label{tab:mia_unlearn}
\end{table*}

\textbf{(4) Knowledge leakage attacks for the unlearning data \(\mathcal{D}_u\).} Here, attackers construct a knowledge leakage model \(K\), which outputs the predictive accuracy on \(\mathcal{D}_t\) as a proxy for retained knowledge after unlearning, where \(\mathcal{D}_t\) is either the unlearning dataset \(\mathcal{D}_u\) or drawn from the same domain \cite{yuan2025towards,doshi2024does}. Depending on the attacking tools adopted to amplify the leakage, such attacks can be further classified into two categories: \emph{perturbing unlearned model} and \emph{perturbing unlearned data}. In the perturbing unlearned model setting, attackers generate a nearby variant \(\widetilde{\theta}_u\) of \(\theta_u\) to use for querying. For example, \cite{doshi2024does} fine-tunes $\theta_u$ with an external dataset \(\mathcal{D}_{ft}\), resulting in a perturbed model \(\theta_{ft}=\widetilde{\theta}_u\). In contrast, in the perturbing unlearned data setting, attackers perturb the unlearned data \(\mathcal{D}_u\) to construct a modified dataset \(\widetilde{\mathcal{D}}_u\) for querying the unlearned model \(\theta_u\) \cite{yuan2025towards}.

\subsection{Benchmark Design for Defenses}

In our benchmark, we also evaluate the state-of-the-art defenses (see Table \ref{tab:UnlPrivacyDefense}), which address unlearning-induced privacy risks. To defend against membership inference attacks on unlearning data \(\mathcal{D}_u\) that exploit discrepancies between the pre-trained model \(\theta\) and unlearned model \(\theta_u\), \cite{wang2025crfu} proposes a defense based on minimizing the mutual information between the learned representation and the unlearning data. To counter privacy attacks that perturb the unlearned data \(\mathcal{D}_u\), \cite{yuan2025towards} introduces adversarial suffix training and then employs latent adversarial unlearning to suppress residual knowledge leakage. \cite{wang2025crfu} also tackles reconstruction attacks targeting \(\mathcal{D}_u\). In response to privacy risks induced by perturbations to the unlearned model \(\theta_u\), \cite{fan2025towards} develops a robust unlearning framework grounded in sharpness-aware minimization (SAM). \cite{tamirisa2025tamperresistant} proposes tampering attack resistance (TAR) that applies tampering attacks on \(\theta\) and adversarial unlearning to improve the robustness of \(\theta_u\).

\section{Experiments}
\label{sec:exp}

Here, we present comprehensive experiments to establish the PrivUB benchmark. \emph{More experimental details and results can be found in the full version of this paper}.

\textbf{Unlearning methods.} In experiments, we adopt popular unlearning methods in deep learning and LLM settings. For the deep learning setting, we use retraining from scratch, SISA~\cite{bourtoule2021machine}, Finetune (FT)~\cite{WarPirWreRie20}, Influence Unlearning (IU)~\cite{izzo2021approximate},  NegGrad+~\cite{kurmanji2023towards}, Gradient Ascent (GA)~\cite{thudi2022unrolling}, SCRUB~\cite{kurmanji2023towards}, and SalUn~\cite{fan2024salun}. For the LLM setting, we adopt the Gradient Ascent (GA)~\cite{yao2024large}, Negative Preference Optimization (NPO)~\cite{zhang2024negative}, and Representation Misdirection for Unlearning (RMU)~\cite{li2024wmdp}. Additionally, due to space limitations, more experiments on uncertainty-aware machine unlearning methods can be found in the full version of the paper.

\textbf{Datasets.} In experiments, we adopt a diverse set of real-world datasets: Chest X-Ray~\cite{kermany2018identifying}, CelebA~\cite{liu2015faceattributes}, CIFAR-10, CIFAR-100~\cite{krizhevsky2009cifar}, WMDP-Biology, WMDP-Cyber~\cite{li2024wmdp}, RWKU~\cite{jin2024rwku}, Openwebtext~\cite{Gokaslan2019OpenWeb}, AG-News~\cite{zhang2015character}, Wikitext-103~\cite{merity2016pointer}, and XSum~\cite{xsum-emnlp}.

\textbf{Models.} In experiments, we consider mainstream models, including ResNet-50~\cite{he2016deep}, VGG-19~\cite{simonyan2014very}, ResNet-18~\cite{he2016deep}, ConvNet, Llama-2-13B~\cite{touvron2023llama}, Llama-3-8B~\cite{grattafiori2024llama}, Llama-2-7B~\cite{touvron2023llama}, Zephyr-7B-beta~\cite{tunstall2023zephyr}, Phi-3~\cite{abdin2024phi}, and GPTNeo-1.3B~\cite{gao2020pile}.

\textbf{Privacy attacks and defenses.} In experiments, we evaluate a range of privacy attacks in selective forgetting, including membership inference attacks~\cite{gu2024auditing,du2024textual,lu2022label,lu2022fp,carlini2022privacy,chen2021machine}, data reconstruction attacks~\cite{wang2025tape,hu2024learn,du2024textual}, and knowledge leakage attacks~\cite{hu2025jogging,zhang2025catastrophic,lucki2025adversarial,xuan2025unlearning,hsu2025are,yuan2025towards,doshi2024does}. We also evaluate the defenses~\cite{fan2025towards,tamirisa2025tamperresistant,yuan2025towards,wang2025crfu} against privacy vulnerabilities in unlearning.

\textbf{Evaluation metrics.} To evaluate privacy leakage, we use a variety of metrics tailored to each attack scenario. For membership inference, we use standard metrics~\cite{carlini2022membership} including MIA accuracy (Acc), AUC, and ROC curve. We also use the failure rate and the empirical CDF for detecting the privacy risks in retain data. For data reconstruction, we measure the data recovery quality using cosine similarity (CS) and mean squared error (MSE)~\cite{hu2024learn}. For knowledge leakage, we adopt unlearning accuracy, test accuracy, MIA Acc, and ROUGE score~\cite{maini2024tofu}.

\begin{table*}[t!]
\small
\centering
\begin{tabular}{@{\hspace{5pt}}c@{\hspace{5pt}}c|c@{\hspace{5pt}}c|c@{\hspace{5pt}}c|c@{\hspace{5pt}}c}
\hline
\multirow{2}{*}{\textbf{\makecell{Model access \\ information}}} 
& \multirow{2}{*}{\textbf{Method}}
& \multicolumn{2}{c|}{\textbf{Chest X-Ray}}
& \multicolumn{2}{c|}{\textbf{CelebA}}
& \multicolumn{2}{c}{\textbf{CIFAR-10}}\\
\cline{3-8}
& & \textbf{MSE} \(\downarrow\) & \textbf{CS} \(\uparrow\) & \textbf{MSE} \(\downarrow\) & \textbf{CS} \(\uparrow\) & \textbf{MSE} \(\downarrow\) & \textbf{CS} \(\uparrow\) \\
\hline
\multirow{3}{*}{\makecell{Model \\ weights}} & Basic DRA & 0.208 $\pm$ 0.004 & 0.499 $\pm$ 0.002 & 0.245 $\pm$ 0.015 & 0.567 $\pm$ 0.003 & 0.265 $\pm$ 0.016 & 0.656 $\pm$ 0.008 \\
& \cite{hu2024learn} & 0.064 $\pm$ 0.003 & 0.897 $\pm$ 0.003 & 0.075 $\pm$ 0.004 & 0.844 $\pm$ 0.015 & 0.062 $\pm$ 0.008  & 0.892 $\pm$ 0.014 \\
& \cite{du2024textual} & 0.030 $\pm$ 0.002 & 0.953 $\pm$ 0.003 & 0.141 $\pm$ 0.008 & 0.772 $\pm$ 0.009 & 0.109 $\pm$ 0.005  & 0.841 $\pm$ 0.008 \\ 
\hline
\multirow{2}{*}{Posterior} & Basic DRA & 0.191 $\pm$ 0.004 & 0.606 $\pm$ 0.005 & 0.248 $\pm$ 0.016 & 0.569 $\pm$ 0.003 & 0.264 $\pm$ 0.016 & 0.638 $\pm$ 0.009 \\
& \cite{wang2025tape} & 0.050 $\pm$ 0.000 & 0.920 $\pm$ 0.004 & 0.038 $\pm$ 0.002 & 0.908 $\pm$ 0.007 & 0.017 $\pm$ 0.001 & 0.970 $\pm$ 0.003 \\
\hline
\end{tabular}
\caption{Comparisons of data reconstruction attacks for unlearning data.}
\label{tab:dr_unlearn}
\end{table*}

\subsection{Experiments on Membership Inference Attacks}
\begin{figure}[htbp]
\centering
\begin{minipage}{0.465\linewidth}
\centering
\includegraphics[width=1\linewidth]{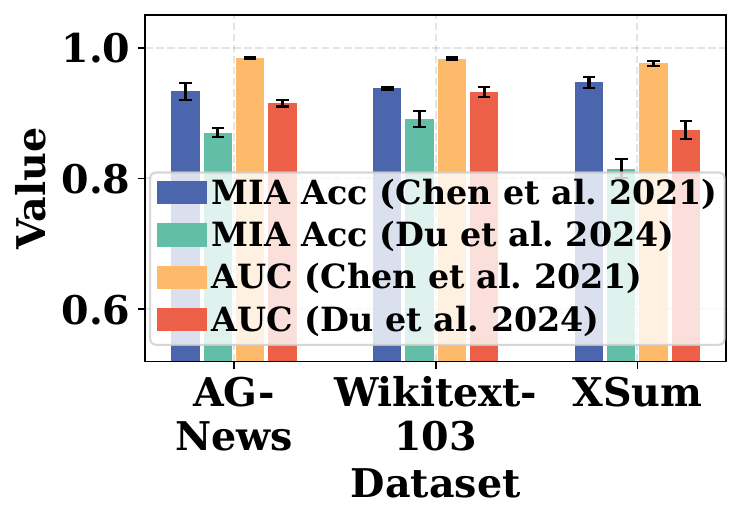}
\caption{Membership inference for unlearning data.}
\label{fig:mia_unlearn_llm}
\end{minipage}
\hfill
\begin{minipage}{.465\linewidth}
\centering
\includegraphics[width=1\textwidth]{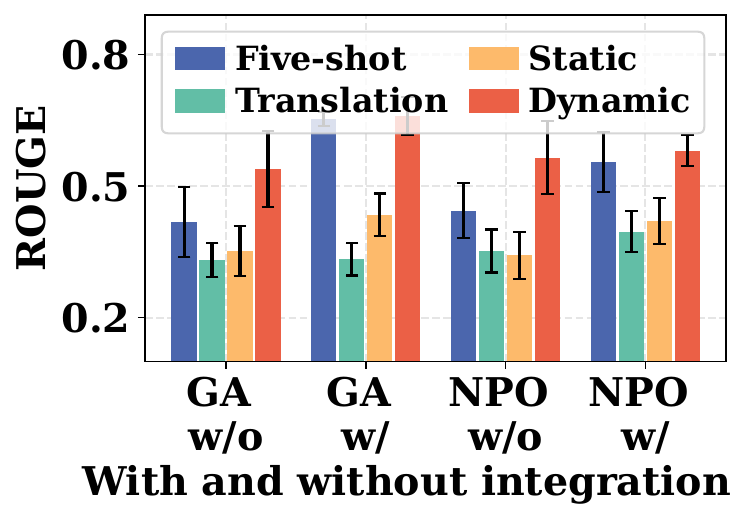}
\caption{Knowledge leakage with perturbing model.}
\label{fig:kl_combine}
\end{minipage}
\end{figure}

First, we investigate the effectiveness of membership inference for unlearning data using pre-trained and unlearned model discrepancy. We categorize the existing approaches~\cite{lu2022label,lu2022fp,chen2021machine} based on the model access information. Table~\ref{tab:mia_unlearn} illustrates the MIA accuracy and AUC across various datasets using ResNet-18 with retraining. We compare these methods with basic MIA baselines that query only the pre-trained model. Fig.~\ref{fig:mia_unlearn_llm} also presents the results of applying~\cite{chen2021machine}, originally designed for deep learning models, to LLMs, and compares with~\cite{du2024textual}. Here, we adopt the GPTNeo-1.3B model. From these results, we have the following observations: (1) The discrepancy between pre-trained and unlearned models reveals unintended information, enabling privacy attacks that surpass classical membership inference on the pre-trained model. (2) For the attack method in~\cite{chen2021machine}, access to richer query information (e.g., from label-only to full posterior (Pos)) leads to improved attack performance. (3) The attack methods in~\cite{lu2022label, lu2022fp}, which leverage adversarial example strategies, exhibit strong performance, as validated in their works. (4) Privacy attacks can be effectively generalized from deep learning models to LLMs, maintaining high performance.

\begin{figure}[htbp]
\centering
\begin{subfigure}{0.495\linewidth}
\includegraphics[width=1\linewidth]{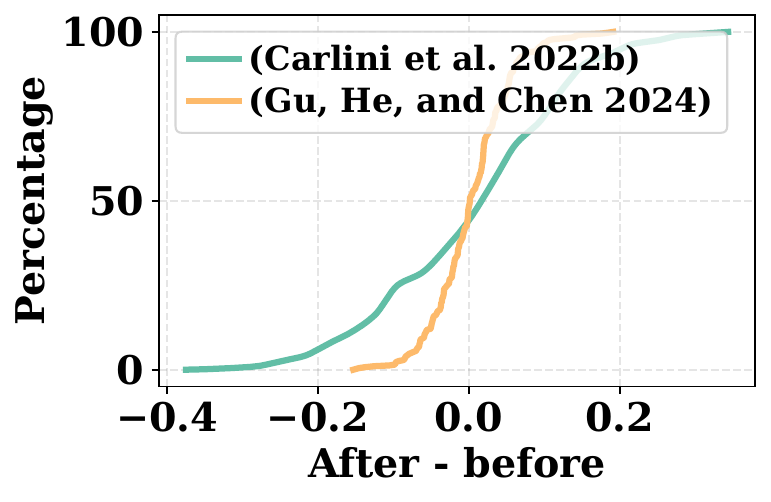}
\caption{Chest X-Ray}
\end{subfigure}
\begin{subfigure}{0.495\linewidth}
\includegraphics[width=1\linewidth]{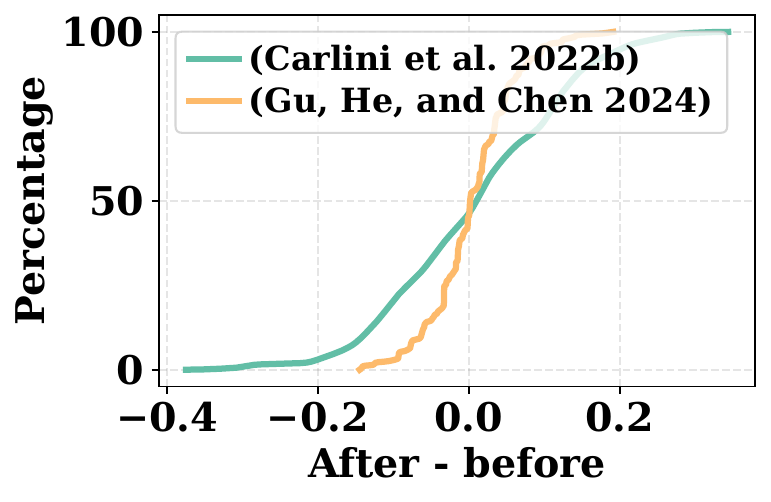}
\caption{CelebA}
\end{subfigure}
\caption{Empirical CDF of membership inference attacks before and after unlearning on retain data.}
\label{fig:mia_retain_emcdf}
\end{figure}

Then, we explore the impact of unlearning on retain data with membership inference attacks. We consider a setting where 10\% of randomly selected training samples are removed using retraining, and then evaluate MIA performance using LiRA~\cite{carlini2022privacy} and A-LiRA~\cite{gu2024auditing}. Fig.~\ref{fig:mia_retain_emcdf} presents the predicted MIA accuracy on the retain set before and after unlearning with ResNet-18. We find the following observations: (1) Privacy of many samples in the retain set deteriorates after unlearning is applied to the forget set, indicating hidden privacy vulnerabilities of unlearning. (2) LiRA generally worsens the privacy of the retain set after unlearning, compared to A-LiRA, which aims to reduce computational overhead in LiRA.

\subsection{Experiments on Data Reconstruction Attacks}

Here, we evaluate the performance of data reconstruction for unlearning data leveraging pre-trained and unlearned model discrepancy. We perform data reconstruction attacks for existing methods~\cite{wang2025tape,hu2024learn,du2024textual}, which aim to recover sensitive data features from unlearned models by retraining. Among these, \cite{du2024textual} is extended from LLMs to deep learning models. In contrast, we employ two baselines that optimize directly against the prediction loss on the target data without access to the unlearned models. Table~\ref{tab:dr_unlearn} shows the results on various datasets using ConvNet. Based on the obtained results,  we observe the following: (1) The discrepancy between the pre-trained and unlearned models reveals significant information about the unlearning samples and enables better reconstruction than using the pre-trained model alone. (2) The posterior augmentation strategy in~\cite{wang2025tape} contributes to its strong reconstruction performance. (3) Privacy attacks originally designed for LLMs can be adapted to deep learning models, achieving competitive performance.

\subsection{Experiments on Knowledge Leakage Attacks}
\begin{figure}[htbp]
\centering
\begin{subfigure}{0.495\linewidth}
\includegraphics[width=1\linewidth]{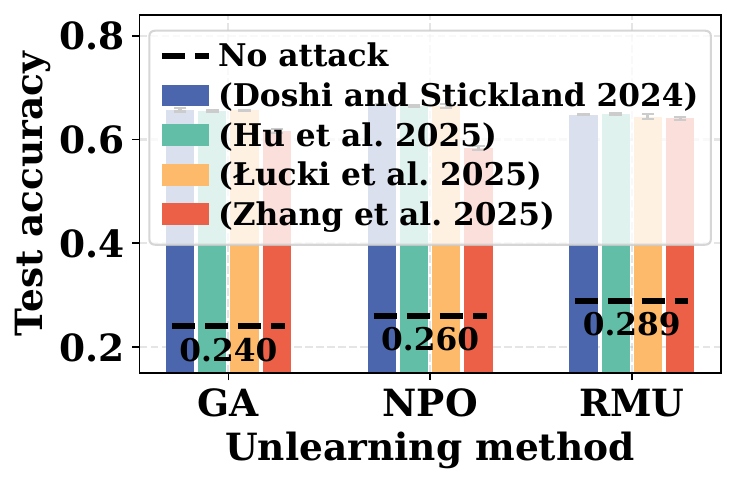}
\caption{WMDP-Biology}
\end{subfigure}
\begin{subfigure}{0.495\linewidth}
\includegraphics[width=1\linewidth]{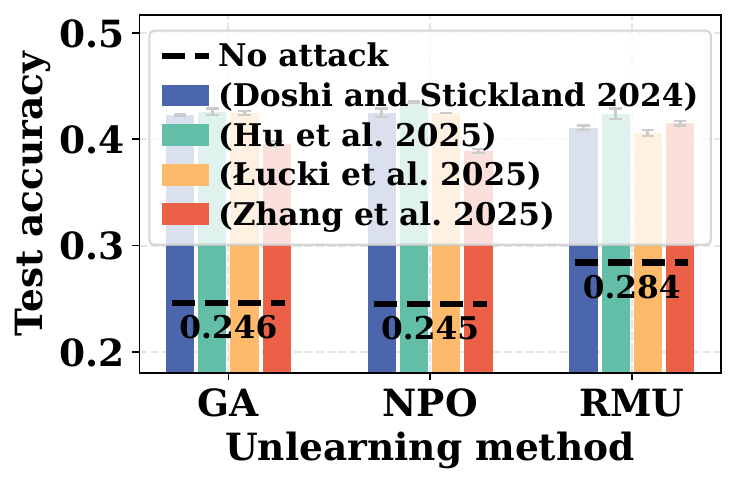}
\caption{WMDP-Cyber}
\end{subfigure}
\caption{Comparisons of knowledge leakage attacks for unlearning data using perturbing unlearned model methods.}
\label{fig:kl_compare_model}
\end{figure}

Here, we explore the performance of knowledge leakage for unlearning data using perturbing unlearned model methods. Specifically, we apply the methods of fine-tuning external data (Openwebtext)~\cite{doshi2024does}, fine-tuning partial unlearning data~\cite{hu2025jogging}, fine-tuning retain data~\cite{lucki2025adversarial}, and using model quantization~\cite{zhang2025catastrophic} to the unlearned model to test the recovered knowledge of the unlearning data. Fig.~\ref{fig:kl_compare_model} presents the test accuracy of the WMDP biology and cybersecurity knowledge recovered by each method on the unlearned model of Zephyr-7B-beta. From these results, we make the following observations: (1) The privacy vulnerabilities of knowledge leakage exist in various selective forgetting methods. (2) Perturbing the model through fine-tuning or quantization can effectively recover unlearned knowledge, with fine-tuning methods generally yielding better performance. (3) Among fine-tuning approaches, using partial unlearning data typically achieves better recovery performance than using external data or retain data.

\begin{figure}[htbp]
\centering
\begin{subfigure}{0.495\linewidth}
\includegraphics[width=1\linewidth]{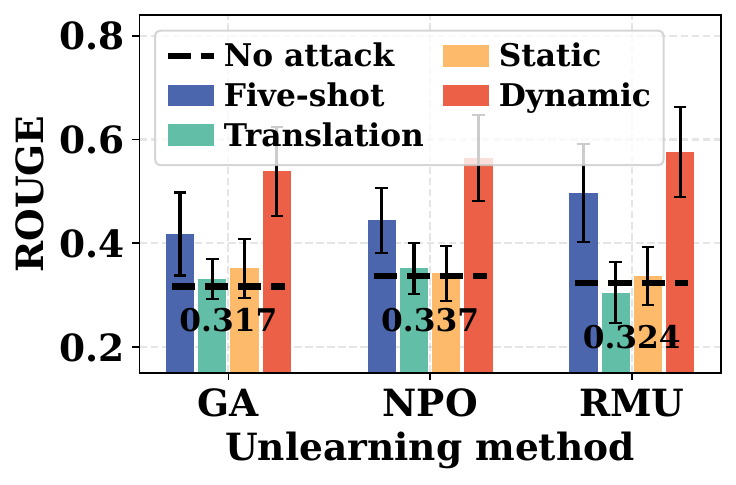}
\caption{LLMs}
\label{fig:kl_unlearn_data_dl_a}
\end{subfigure}
\begin{subfigure}{0.495\linewidth}
\includegraphics[width=1\linewidth]{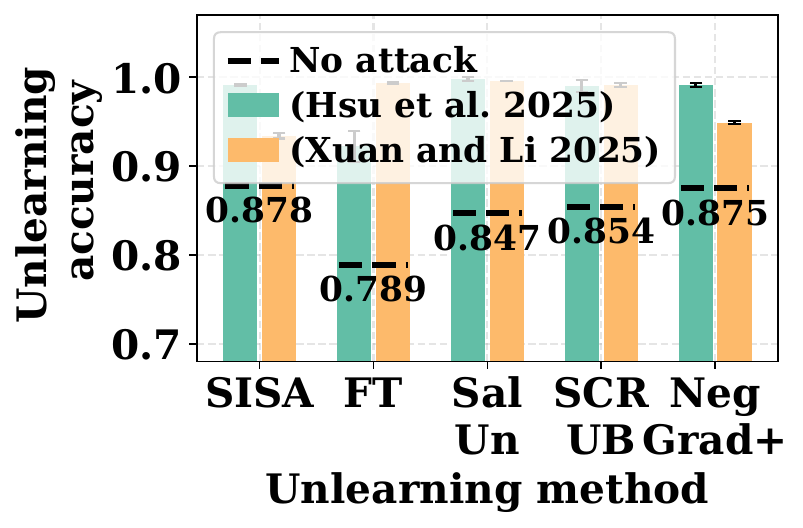}
\caption{Deep learning models}
\label{fig:kl_unlearn_data_dl_b}
\end{subfigure}
\caption{Comparisons of knowledge leakage attacks for unlearning data via perturbing unlearned data methods.}
\label{fig:kl_unlearn_data_dl}
\end{figure}

We also examine the performance of knowledge leakage on unlearning data via perturbing unlearned data methods. Specifically, in the LLM setting, we apply the prompt perturbation strategies, including five-shot prompting and translation~\cite{doshi2024does}, and static prefix injection and the dynamic adversarial suffix optimization~\cite{yuan2025towards}.  In the deep learning setting, we compare image perturbations using adversarial examples generated by FGSM~\cite{hsu2025are} and the gradient-based optimization~\cite {xuan2025unlearning}. Fig.~\ref{fig:kl_unlearn_data_dl_a} shows the ROUGE score of unlearned knowledge on the RWKU dataset using Llama-3-8B. Fig.~\ref{fig:kl_unlearn_data_dl_b} presents the unlearning data accuracy under a perturbation size of 8/255 on CIFAR-10 with ResNet-18. Based on these results, we find the following observations: (1) Data perturbations can substantially increase the privacy risks of unlearning data in both LLMs and deep learning models. (2) Optimization-based approaches that aim to recover correct outputs tend to outperform static methods in revealing residual knowledge for unlearning data.

\begin{figure*}[t!]
\centering
\centering
\begin{subfigure}{0.245\linewidth}
\includegraphics[width=1\linewidth]{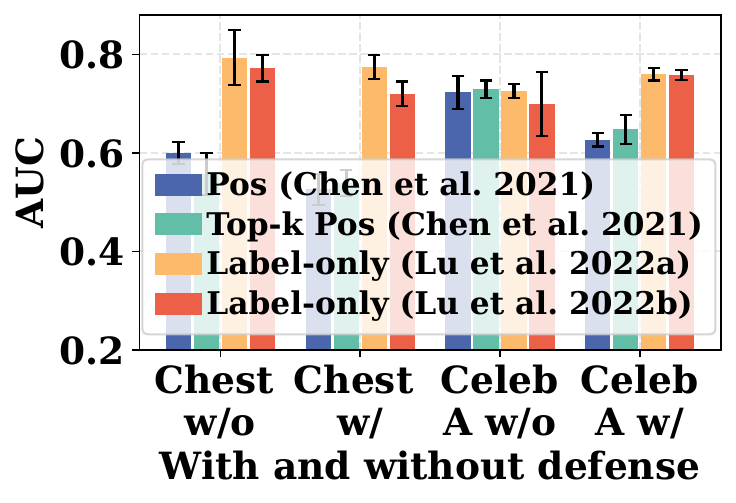}
\caption{Membership inference}
\label{fig:defense_a}
\end{subfigure}
\begin{subfigure}{0.245\linewidth}
\includegraphics[width=1\linewidth]{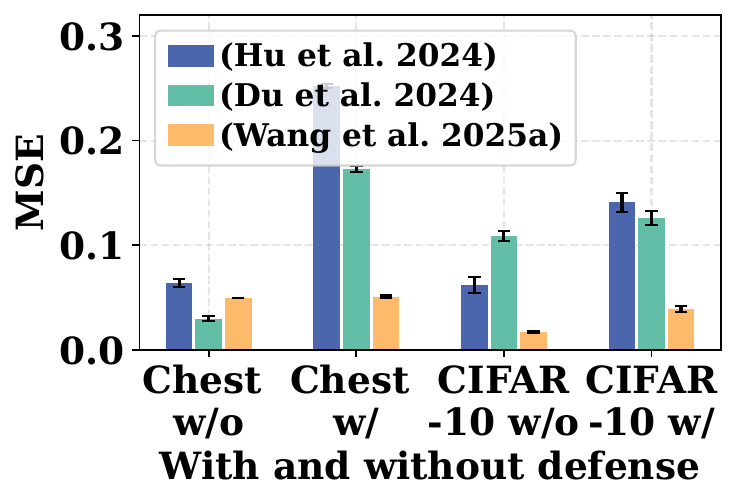}
\caption{Data reconstruction}
\label{fig:defense_b}
\end{subfigure}
\begin{subfigure}{0.245\linewidth}
\includegraphics[width=1\linewidth]{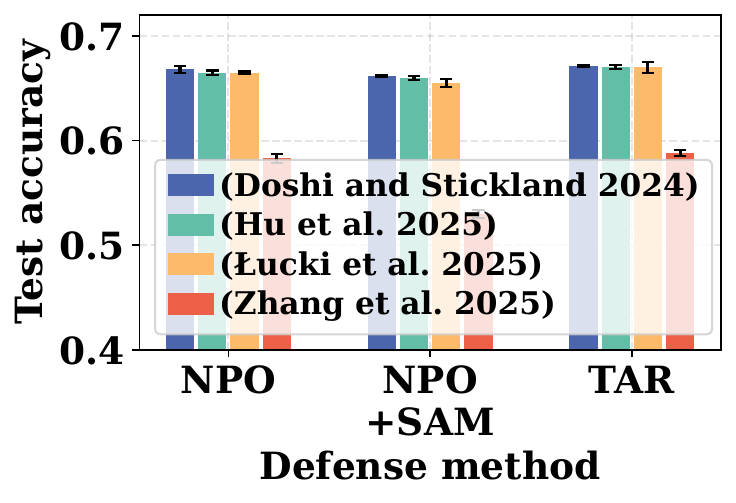}
\caption{Perturbing unlearned model}
\label{fig:defense_c}
\end{subfigure}
\begin{subfigure}{0.245\linewidth}
\includegraphics[width=1\linewidth]{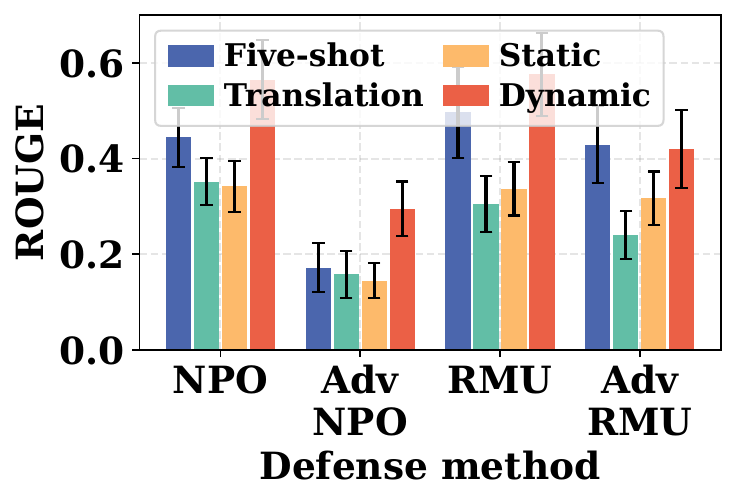}
\caption{Perturbing unlearned data}
\label{fig:defense_d}
\end{subfigure}
\caption{Defenses against privacy attacks in unlearning.}
\label{fig:defense}
\end{figure*}

To further evaluate the impact of privacy vulnerabilities in unlearning, we combine the attacking tools in knowledge leakage and investigate their coordinated effects.  Fig.~\ref{fig:kl_combine} presents the knowledge leakage for unlearning data using various perturbing unlearned data methods, integrated with the perturbing unlearned model approach from~\cite{hu2025jogging}, which fine-tunes partial unlearning data. Notably, the attack performance of each perturbing unlearned data method increases after integration. From these results, we observe that combining multiple attacking tools within the same vulnerability type can improve the attack effectiveness and lead to greater privacy leakage in selective forgetting.

Additionally, we explore the impact of combining attacking tools across different types of privacy vulnerabilities. Fig.~\ref{fig:mia_combine} shows the membership inference results of using pre-trained and unlearned model discrepancy with and without the perturbing unlearned data method in knowledge leakage. Specifically, we apply PGD-based perturbations following~\cite{hsu2025are} and conduct the label-only membership inference attacks. We find that the MIA accuracy is significantly boosted after applying the data perturbations. From these results, we observe that different types of privacy vulnerabilities can be integrated to exacerbate the privacy risks.

\subsection{Experiments on Defenses Against Privacy Attacks}

Here, we assess existing defense mechanisms designed to mitigate the information leakage in unlearning. First, for attacks that exploit discrepancies between the pre-trained and unlearned models, we leverage the representation compression method~\cite{wang2025crfu} to defend against membership inference and data reconstruction, with the results reported in Fig.~\ref{fig:defense_a} and Fig.~\ref{fig:defense_b}. Then, we adopt the robust unlearning method NPO+SAM~\cite{fan2025towards} and TAR~\cite{tamirisa2025tamperresistant}, aiming to defend against knowledge leakage from perturbing unlearned models. The corresponding results are presented in Fig.~\ref{fig:defense_c}. Next, we apply the adversarial unlearning method (AdvNPO and AdvRMU) to enhance the unlearning robustness specific to the knowledge leakage from perturbing unlearned data. The results are shown in Fig.~\ref{fig:defense_d}. Based on these defense evaluations, we conclude the following observations: (1) Existing defense mechanisms show limited effectiveness in mitigating privacy leakage in unlearning. (2) The difficulty of defense against privacy vulnerabilities varies across attack types; in particular, defending against perturbing data attacks appears to be more tractable than perturbing model attacks.

\begin{figure}[htbp]
\centering
\begin{minipage}{0.465\linewidth}
\centering
\includegraphics[width=1\linewidth]{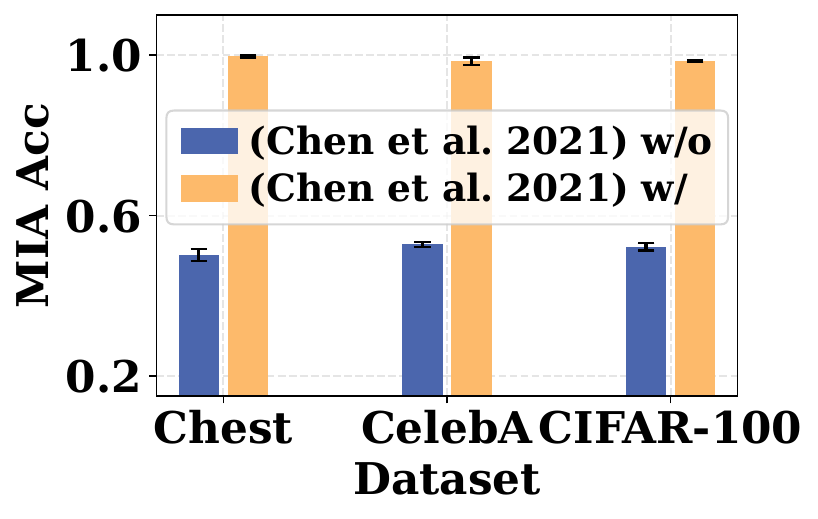}
\caption{MIAs with perturbing unlearned data.}
\label{fig:mia_combine}
\end{minipage}
\hfill
\begin{minipage}{.465\linewidth}
\centering
\includegraphics[width=1\textwidth]{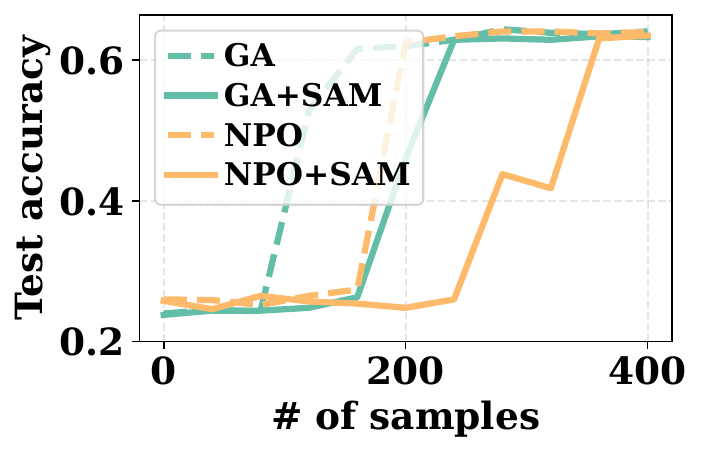}
\caption{Impact of attack samples on defenses.}
\label{fig:impact_number}
\end{minipage}
\end{figure}

Additionally, we examine the robust unlearning~\cite{fan2025towards} with fine-tuning of partial unlearning data~\cite{hu2025jogging} to better understand the limitations of current defense mechanisms. Fig.~\ref{fig:impact_number} presents the test accuracy under varying attack samples on the WMDP-Biology dataset. The results indicate that while SAM shows some resistance to the attacks when the number of attack samples is small, its effectiveness significantly degrades as the number of attack samples increases. These findings suggest that existing defenses are highly sensitive to the attack configurations and often fail to maintain robustness under certain conditions.

\section{Conclusion and Future Work}
\label{sec:conclusion}

In this study, we propose PrivUB, the first comprehensive benchmark for evaluating privacy vulnerabilities in selective forgetting. Our benchmark focuses on three critical dimensions of privacy vulnerabilities: membership inference, data reconstruction, and knowledge leakage, and two categories of victim data: unlearning data and retain data, during the unlearning and deployment phases. We apply PrivUB to systematically evaluate 21 state-of-the-art privacy attacks and defenses under 10 unlearning methods, covering 11 widely-used datasets, 10 representative model architectures, and 10 evaluation metrics. To the best of our knowledge, this is the first work to comprehensively benchmark the privacy vulnerabilities arising from unlearning-induced attacks and their corresponding defenses. Our findings reveal significant privacy risks exposed in current selective forgetting techniques and underscore the need for advanced defenses for future research. These include developing robust unlearning to mitigate privacy leakage both during unlearning and after model deployment. We believe that PrivUB will benefit the community by providing a standardized tool and facilitating faithful privacy assessments.

\section{Acknowledgments}
This work is supported in part by the US National Science Foundation under grants CNS-2350332 and IIS-2442750. Any opinions, findings, and conclusions or recommendations expressed in this material are those of the author(s) and do not necessarily reflect the views of the National Science Foundation.

\bibliography{aaai2026}

@inproceedings{chen2021machine,
  title={When machine unlearning jeopardizes privacy},
  author={Chen, Min and Zhang, Zhikun and Wang, Tianhao and Backes, Michael and Humbert, Mathias and Zhang, Yang},
  booktitle={Proceedings of the 2021 ACM SIGSAC conference on computer and communications security},
  pages={896--911},
  year={2021}
}

@article{lu2022label,
  title={Label-only membership inference attacks on machine unlearning without dependence of posteriors},
  author={Lu, Zhaobo and Liang, Hai and Zhao, Minghao and Lv, Qingzhe and Liang, Tiancai and Wang, Yilei},
  journal={International Journal of Intelligent Systems},
  volume={37},
  number={11},
  pages={9424--9441},
  year={2022},
  publisher={Wiley Online Library}
}

@inproceedings{lu2022fp,
  title={Fp 2-mia: A membership inference attack free of posterior probability in machine unlearning},
  author={Lu, Zhaobo and Wang, Yilei and Lv, Qingzhe and Zhao, Minghao and Liang, Tiancai},
  booktitle={International Conference on Provable Security},
  pages={167--175},
  year={2022},
  organization={Springer}
}

@article{du2024textual,
  title={Textual unlearning gives a false sense of unlearning},
  author={Du, Jiacheng and Wang, Zhibo and Zhang, Jie and Pang, Xiaoyi and Hu, Jiahui and Ren, Kui},
  journal={arXiv preprint arXiv:2406.13348},
  year={2024}
}

@inproceedings{hu2024learn,
  title={Learn what you want to unlearn: Unlearning inversion attacks against machine unlearning},
  author={Hu, Hongsheng and Wang, Shuo and Dong, Tian and Xue, Minhui},
  booktitle={2024 IEEE Symposium on Security and Privacy (SP)},
  pages={3257--3275},
  year={2024},
  organization={IEEE}
}

@inproceedings{wang2025tape,
  title={TAPE: Tailored Posterior Difference for Auditing of Machine Unlearning},
  author={Wang, Weiqi and Tian, Zhiyi and Liu, An and Yu, Shui},
  booktitle={Proceedings of the ACM on Web Conference 2025},
  pages={3061--3072},
  year={2025}
}

@article{carlini2022privacy,
  title={The privacy onion effect: Memorization is relative},
  author={Carlini, Nicholas and Jagielski, Matthew and Zhang, Chiyuan and Papernot, Nicolas and Terzis, Andreas and Tramer, Florian},
  journal={Advances in Neural Information Processing Systems},
  volume={35},
  pages={13263--13276},
  year={2022}
}

@misc{
gu2024auditing,
title={Auditing Privacy Protection of Machine Unlearning},
author={Yuechun Gu and Jiajie He and Keke Chen},
year={2024},
url={https://openreview.net/forum?id=Uv7bWrIucU}
}

@misc{
xuan2025unlearning,
title={Unlearning Mapping Attack: Exposing Hidden Vulnerabilities in Machine Unlearning},
author={Hao Xuan and Xingyu Li},
year={2025},
url={https://openreview.net/forum?id=KvFk356RpR}
}

@inproceedings{
hsu2025are,
title={Are We Really Unlearning? The Presence of Residual Knowledge in Machine Unlearning},
author={Hsiang Hsu and Pradeep Niroula and Zichang He and Chun-Fu Chen},
booktitle={I Can't Believe It's Not Better: Challenges in Applied Deep Learning},
year={2025},
url={https://openreview.net/forum?id=HsjHGNYv2O}
}

@inproceedings{yuan2025towards,
  title={Towards robust knowledge unlearning: An adversarial framework for assessing and improving unlearning robustness in large language models},
  author={Yuan, Hongbang and Jin, Zhuoran and Cao, Pengfei and Chen, Yubo and Liu, Kang and Zhao, Jun},
  booktitle={Proceedings of the AAAI Conference on Artificial Intelligence},
  volume={39},
  pages={25769--25777},
  year={2025}
}

@article{doshi2024does,
  title={Does unlearning truly unlearn? a black box evaluation of llm unlearning methods},
  author={Doshi, Jai and Stickland, Asa Cooper},
  journal={arXiv preprint arXiv:2411.12103},
  year={2024}
}

@article{hu2025jogging,
  title={Jogging the Memory of Unlearned LLMs Through Targeted Relearning Attacks},
  author={Hu, Shengyuan and Fu, Yiwei and Wu, Zhiwei Steven and Smith, Virginia},
  journal={International Conference on Learning Representations},
  year={2025}
}

@article{zhang2025catastrophic,
  title={Catastrophic Failure of LLM Unlearning via Quantization},
  author={Zhang, Zhiwei and Wang, Fali and Li, Xiaomin and Wu, Zongyu and Tang, Xianfeng and Liu, Hui and He, Qi and Yin, Wenpeng and Wang, Suhang},
  journal={International Conference on Learning Representations},
  year={2025}
}

@article{lucki2025adversarial,
  title={An adversarial perspective on machine unlearning for ai safety},
  author={{\L}ucki, Jakub and Wei, Boyi and Huang, Yangsibo and Henderson, Peter and Tram{\`e}r, Florian and Rando, Javier},
  journal={Transactions on Machine Learning Research},
  year={2025}
}

@article{fan2025towards,
  title={Towards llm unlearning resilient to relearning attacks: A sharpness-aware minimization perspective and beyond},
  author={Fan, Chongyu and Jia, Jinghan and Zhang, Yihua and Ramakrishna, Anil and Hong, Mingyi and Liu, Sijia},
  journal={International conference on machine learning},
  year={2025}
}

@inproceedings{
tamirisa2025tamperresistant,
title={Tamper-Resistant Safeguards for Open-Weight {LLM}s},
author={Rishub Tamirisa and Bhrugu Bharathi and Long Phan and Andy Zhou and Alice Gatti and Tarun Suresh and Maxwell Lin and Justin Wang and Rowan Wang and Ron Arel and Andy Zou and Dawn Song and Bo Li and Dan Hendrycks and Mantas Mazeika},
booktitle={The Thirteenth International Conference on Learning Representations},
year={2025},
}

@article{wang2025crfu,
  title={CRFU: Compressive Representation Forgetting Against Privacy Leakage on Machine Unlearning},
  author={Wang, Weiqi and Zhang, Chenhan and Tian, Zhiyi and Liu, Shushu and Yu, Shui},
  journal={IEEE Transactions on Dependable and Secure Computing},
  year={2025},
  publisher={IEEE}
}

@article{kermany2018identifying,
title = {Identifying Medical Diagnoses and Treatable Diseases by Image-Based Deep Learning},
journal = {Cell},
year = {2018},
author = {Daniel S. Kermany and Michael Goldbaum and Wenjia Cai and Carolina C.S. Valentim and Huiying Liang and Sally L. Baxter and Alex McKeown and Ge Yang and Xiaokang Wu and Fangbing Yan and Justin Dong and Made K. Prasadha and Jacqueline Pei and Magdalene Y.L. Ting and Jie Zhu and Christina Li and Sierra Hewett and Jason Dong and Ian Ziyar and Alexander Shi and Runze Zhang and Lianghong Zheng and Rui Hou and William Shi and Xin Fu and Yaou Duan and Viet A.N. Huu and Cindy Wen and Edward D. Zhang and Charlotte L. Zhang and Oulan Li and Xiaobo Wang and Michael A. Singer and Xiaodong Sun and Jie Xu and Ali Tafreshi and M. Anthony Lewis and Huimin Xia and Kang Zhang},
}

@inproceedings{liu2015faceattributes,
  title = {Deep Learning Face Attributes in the Wild},
  author = {Liu, Ziwei and Luo, Ping and Wang, Xiaogang and Tang, Xiaoou},
  booktitle = {Proceedings of International Conference on Computer Vision (ICCV)},
  month = {December},
  year = {2015} 
}

@article{krizhevsky2009cifar,
  title={Cifar-10 and cifar-100 datasets},
  author={Krizhevsky, Alex and Nair, Vinod and Hinton, Geoffrey},
  journal={URl: https://www. cs. toronto. edu/kriz/cifar. html},
  volume={6},
  number={1},
  pages={1},
  year={2009},
  note= {Accessed: July 2025}
}

@article{li2024wmdp,
  title={The wmdp benchmark: Measuring and reducing malicious use with unlearning},
  author={Li, Nathaniel and Pan, Alexander and Gopal, Anjali and Yue, Summer and Berrios, Daniel and Gatti, Alice and Li, Justin D and Dombrowski, Ann-Kathrin and Goel, Shashwat and Phan, Long and others},
  journal={Proceedings of the 41st International Conference on Machine Learning},
  year={2024}
}

@inproceedings{
jin2024rwku,
title={{RWKU}: Benchmarking Real-World Knowledge Unlearning for Large Language Models},
author={Zhuoran Jin and Pengfei Cao and Chenhao Wang and Zhitao He and Hongbang Yuan and Jiachun Li and Yubo Chen and Kang Liu and Jun Zhao},
booktitle={The Thirty-eight Conference on Neural Information Processing Systems Datasets and Benchmarks Track},
year={2024}
}

@misc{Gokaslan2019OpenWeb,
    title={OpenWebText Corpus},
    author={Gokaslan, Aaron and Cohen, Vanya and Pavlick, Ellie and Tellex, Stefanie},
    howpublished={\url{http://Skylion007.github.io/OpenWebTextCorpus}},
    year={2019},
    note= {Accessed: July 2025}
}

@article{zhang2015character,
  title={Character-level convolutional networks for text classification},
  author={Zhang, Xiang and Zhao, Junbo and LeCun, Yann},
  journal={Advances in neural information processing systems},
  volume={28},
  year={2015}
}

@misc{merity2016pointer,
      title={Pointer Sentinel Mixture Models},
      author={Stephen Merity and Caiming Xiong and James Bradbury and Richard Socher},
      year={2016},
      eprint={1609.07843},
      archivePrefix={arXiv},
      primaryClass={cs.CL}
}

@InProceedings{xsum-emnlp,
  author =      "Shashi Narayan and Shay B. Cohen and Mirella Lapata",
  title =       "Don't Give Me the Details, Just the Summary! {T}opic-Aware Convolutional Neural Networks for Extreme Summarization",
  booktitle =   "Proceedings of the 2018 Conference on Empirical Methods in Natural Language Processing ",
  year =        "2018",
  address =     "Brussels, Belgium",
}

@inproceedings{he2016deep,
  title={Deep residual learning for image recognition},
  author={He, Kaiming and Zhang, Xiangyu and Ren, Shaoqing and Sun, Jian},
  booktitle={Proceedings of the IEEE conference on computer vision and pattern recognition},
  pages={770--778},
  year={2016}
}

@article{grattafiori2024llama,
  title={The llama 3 herd of models},
  author={Grattafiori, Aaron and Dubey, Abhimanyu and Jauhri, Abhinav and Pandey, Abhinav and Kadian, Abhishek and Al-Dahle, Ahmad and Letman, Aiesha and Mathur, Akhil and Schelten, Alan and Vaughan, Alex and others},
  journal={arXiv preprint arXiv:2407.21783},
  year={2024}
}

@article{touvron2023llama,
  title={Llama 2: Open foundation and fine-tuned chat models},
  author={Touvron, Hugo and Martin, Louis and Stone, Kevin and Albert, Peter and Almahairi, Amjad and Babaei, Yasmine and Bashlykov, Nikolay and Batra, Soumya and Bhargava, Prajjwal and Bhosale, Shruti and others},
  journal={arXiv preprint arXiv:2307.09288},
  year={2023}
}

@article{tunstall2023zephyr,
  title={Zephyr: Direct distillation of lm alignment},
  author={Tunstall, Lewis and Beeching, Edward and Lambert, Nathan and Rajani, Nazneen and Rasul, Kashif and Belkada, Younes and Huang, Shengyi and Von Werra, Leandro and Fourrier, Cl{\'e}mentine and Habib, Nathan and others},
  journal={arXiv preprint arXiv:2310.16944},
  year={2023}
}

@article{gao2020pile,
  title={The Pile: An 800GB Dataset of Diverse Text for Language Modeling},
  author={Gao, Leo and Biderman, Stella and Black, Sid and Golding, Laurence and Hoppe, Travis and Foster, Charles and Phang, Jason and He, Horace and Thite, Anish and Nabeshima, Noa and others},
  journal={arXiv preprint arXiv:2101.00027},
  year={2020}
}

@inproceedings{WarPirWreRie20,
    title={Machine Unlearning of Features and Labels},
    author={Alexander Warnecke and Lukas Pirch and Christian Wressnegger and Konrad Rieck},
    year={2023},
    booktitle={Proc. of the 30th Network and Distributed System Security (NDSS)}
  }

@inproceedings{izzo2021approximate,
  title={Approximate data deletion from machine learning models},
  author={Izzo, Zachary and Smart, Mary Anne and Chaudhuri, Kamalika and Zou, James},
  booktitle={International conference on artificial intelligence and statistics},
  pages={2008--2016},
  year={2021},
  organization={PMLR}
}

@article{kurmanji2023towards,
  title={Towards unbounded machine unlearning},
  author={Kurmanji, Meghdad and Triantafillou, Peter and Hayes, Jamie and Triantafillou, Eleni},
  journal={Advances in neural information processing systems},
  volume={36},
  pages={1957--1987},
  year={2023}
}

@inproceedings{
fan2024salun,
title={SalUn: Empowering Machine Unlearning via Gradient-based Weight Saliency in Both Image Classification and Generation},
author={Chongyu Fan and Jiancheng Liu and Yihua Zhang and Eric Wong and Dennis Wei and Sijia Liu},
booktitle={The Twelfth International Conference on Learning Representations},
year={2024},
url={https://openreview.net/forum?id=gn0mIhQGNM}
}

@inproceedings{
zhang2024negative,
title={Negative Preference Optimization: From Catastrophic Collapse to Effective Unlearning},
author={Ruiqi Zhang and Licong Lin and Yu Bai and Song Mei},
booktitle={First Conference on Language Modeling},
year={2024},
url={https://openreview.net/forum?id=MXLBXjQkmb}
}

@inproceedings{thudi2022unrolling,
  title={Unrolling sgd: Understanding factors influencing machine unlearning},
  author={Thudi, Anvith and Deza, Gabriel and Chandrasekaran, Varun and Papernot, Nicolas},
  booktitle={2022 IEEE 7th European Symposium on Security and Privacy (EuroS\&P)},
  pages={303--319},
  year={2022},
  organization={IEEE}
}

@article{yao2024large,
  title={Large language model unlearning},
  author={Yao, Yuanshun and Xu, Xiaojun and Liu, Yang},
  journal={Advances in Neural Information Processing Systems},
  volume={37},
  pages={105425--105475},
  year={2024}
}

@inproceedings{
maini2024tofu,
title={{TOFU}: A Task of Fictitious Unlearning for {LLM}s},
author={Pratyush Maini and Zhili Feng and Avi Schwarzschild and Zachary Chase Lipton and J Zico Kolter},
booktitle={First Conference on Language Modeling},
year={2024},
url={https://openreview.net/forum?id=B41hNBoWLo}
}

@inproceedings{carlini2022membership,
  title={Membership inference attacks from first principles},
  author={Carlini, Nicholas and Chien, Steve and Nasr, Milad and Song, Shuang and Terzis, Andreas and Tramer, Florian},
  booktitle={2022 IEEE symposium on security and privacy (SP)},
  pages={1897--1914},
  year={2022},
  organization={IEEE}
}

@inproceedings{rombach2022high,
  title={High-resolution image synthesis with latent diffusion models},
  author={Rombach, Robin and Blattmann, Andreas and Lorenz, Dominik and Esser, Patrick and Ommer, Bj{\"o}rn},
  booktitle={Proceedings of the IEEE/CVF conference on computer vision and pattern recognition},
  pages={10684--10695},
  year={2022}
}

@article{hadero2023new,
  title={New York Times sues Microsoft, Open AI over use of content.},
  author={Hadero, Haleluya and Bauder, David},
  journal={Globe \& Mail (Toronto, Canada)},
  pages={B1--B1},
  year={2023},
  publisher={The Globe and Mail Inc.}
}

@inproceedings{yuan2023pseudo,
  title={Pseudo label-guided model inversion attack via conditional generative adversarial network},
  author={Yuan, Xiaojian and Chen, Kejiang and Zhang, Jie and Zhang, Weiming and Yu, Nenghai and Zhang, Yang},
  booktitle={Proceedings of the AAAI Conference on Artificial Intelligence},
  volume={37},
  pages={3349--3357},
  year={2023}
}

@article{li2023privlm,
  title={Privlm-bench: A multi-level privacy evaluation benchmark for language models},
  author={Li, Haoran and Guo, Dadi and Li, Donghao and Fan, Wei and Hu, Qi and Liu, Xin and Chan, Chunkit and Yao, Duanyi and Yao, Yuan and Song, Yangqiu},
  journal={Proceedings of the 62nd Annual Meeting of the Association for Computational Linguistics},
  year={2023}
}

@article{wen2025sok,
  title={SoK: Data Reconstruction Attacks Against Machine Learning Models: Definition, Metrics, and Benchmark},
  author={Wen, Rui and Liu, Yiyong and Backes, Michael and Zhang, Yang},
  journal={USENIX Security Symposium},
  year={2025}
}

@inproceedings{
zhu2024privauditor,
title={PrivAuditor: Benchmarking Data Protection Vulnerabilities in {LLM} Adaptation Techniques},
author={Derui Zhu and Dingfan Chen and Xiongfei Wu and Jiahui Geng and Zhuo Li and Jens Grossklags and Lei Ma},
booktitle={The Thirty-eight Conference on Neural Information Processing Systems Datasets and Benchmarks Track},
year={2024},
url={https://openreview.net/forum?id=VpkfxuVXwx}
}

@inproceedings{song2021systematic,
  title={Systematic evaluation of privacy risks of machine learning models},
  author={Song, Liwei and Mittal, Prateek},
  booktitle={30th USENIX security symposium (USENIX security 21)},
  pages={2615--2632},
  year={2021}
}

@article{niu2025comparing,
  title={Comparing Different Membership Inference Attacks with a Comprehensive Benchmark},
  author={Niu, Jun and Zhu, Xiaoyan and Zeng, Moxuan and Zhang, Ge and Zhao, Qingyang and Huang, Chunhui and Zhang, Yangming and An, Suyu and Wang, Yangzhong and Yue, Xinghui and others},
  journal={IEEE Transactions on Information Forensics and Security},
  year={2025},
  publisher={IEEE}
}

@article{bu2024pre,
  title={Pre-training differentially private models with limited public data},
  author={Bu, Zhiqi and Zhang, Xinwei and Zha, Sheng and Hong, Mingyi and Karypis, George},
  journal={Advances in Neural Information Processing Systems},
  volume={37},
  pages={94652--94683},
  year={2024}
}

@inproceedings{mueller2024llms,
  title={LLMs and memorization: On quality and specificity of copyright compliance},
  author={Mueller, Felix B and G{\"o}rge, Rebekka and Bernzen, Anna K and Pirk, Janna C and Poretschkin, Maximilian},
  booktitle={Proceedings of the AAAI/ACM Conference on AI, Ethics, and Society},
  volume={7},
  pages={984--996},
  year={2024}
}

@article{Chu_Song_Yang_2024, 
title={How to Protect Copyright Data in Optimization of Large Language Models?},
journal={Proceedings of the AAAI Conference on Artificial Intelligence}, 
author={Chu, Timothy and Song, Zhao and Yang, Chiwun}, 
year={2024}
}

@article{regulation2018general,
  title={General data protection regulation},
  author={Regulation, Protection},
  journal={Intouch},
  volume={25},
  pages={1--5},
  year={2018}
}

@book{ccpa2019,
 ISBN = {9781787781320},
 URL = {http://www.jstor.org/stable/j.ctvjghvnn},
 author = {Preston Bukaty},
 publisher = {IT Governance Publishing},
 title = {The California Consumer Privacy Act (CCPA): An implementation guide},
 urldate = {2025-11-08},
 year = {2019}
}

@article{li2025machine,
  title={Machine unlearning: Taxonomy, metrics, applications, challenges, and prospects},
  author={Li, Na and Zhou, Chunyi and Gao, Yansong and Chen, Hui and Zhang, Zhi and Kuang, Boyu and Fu, Anmin},
  journal={IEEE Transactions on Neural Networks and Learning Systems},
  year={2025},
  publisher={IEEE}
}

@article{simonyan2014very,
  title={Very deep convolutional networks for large-scale image recognition},
  author={Simonyan, Karen and Zisserman, Andrew},
  journal={arXiv preprint arXiv:1409.1556},
  year={2014}
}

@techreport{abdin2024phi,
author = {Abdin, Marah I and Ade Jacobs, Sam and Awan, Ammar Ahmad and Aneja, Jyoti and Awadallah, Ahmed and others},
title = {Phi-3 Technical Report: A Highly Capable Language Model Locally on Your Phone},
institution = {Microsoft},
year = {2024},
}

@article{guo2019certified,
  title={Certified data removal from machine learning models},
  author={Guo, Chuan and Goldstein, Tom and Hannun, Awni and Van Der Maaten, Laurens},
  journal={arXiv preprint arXiv:1911.03030},
  year={2019}
}

@inproceedings{bourtoule2021machine,
  title={Machine unlearning},
  author={Bourtoule, Lucas and Chandrasekaran, Varun and Choquette-Choo, Christopher A and Jia, Hengrui and Travers, Adelin and Zhang, Baiwu and Lie, David and Papernot, Nicolas},
  booktitle={2021 IEEE symposium on security and privacy (SP)},
  pages={141--159},
  year={2021},
  organization={IEEE}
}

@article{hu2022lora,
  title={Lora: Low-rank adaptation of large language models.},
  author={Hu, Edward J and Shen, Yelong and Wallis, Phillip and Allen-Zhu, Zeyuan and Li, Yuanzhi and Wang, Shean and Wang, Lu and Chen, Weizhu and others},
  journal={ICLR},
  volume={1},
  number={2},
  pages={3},
  year={2022}
}

@inproceedings{qian2022patient,
  title={Patient similarity learning with selective forgetting},
  author={Qian, Wei and Zhao, Chenxu and Shao, Huajie and Chen, Minghan and Wang, Fei and Huai, Mengdi},
  booktitle={2022 IEEE International Conference on Bioinformatics and Biomedicine (BIBM)},
  pages={529--534},
  year={2022},
  organization={IEEE}
}

@inproceedings{qian2023towards,
  title={Towards understanding and enhancing robustness of deep learning models against malicious unlearning attacks},
  author={Qian, Wei and Zhao, Chenxu and Le, Wei and Ma, Meiyi and Huai, Mengdi},
  booktitle={Proceedings of the 29th ACM SIGKDD Conference on Knowledge Discovery and Data Mining},
  pages={1932--1942},
  year={2023}
}

@article{zhao2023static,
  title={Static and sequential malicious attacks in the context of selective forgetting},
  author={Zhao, Chenxu and Qian, Wei and Ying, Rex and Huai, Mengdi},
  journal={Advances in Neural Information Processing Systems},
  volume={36},
  pages={74966--74979},
  year={2023}
}

@misc{chen2025survey,
  title={A survey of security and privacy issues of machine unlearning},
  author={Chen, Aobo and Li, Yangyi and Zhao, Chenxu and Huai, Mengdi},
  year={2025},
  publisher={Wiley Online Library}
}

@inproceedings{zhao2025membership,
  title={Membership inference attacks with false discovery rate control},
  author={Zhao, Chenxu and Qian, Wei and Chen, Aobo and Huai, Mengdi},
  booktitle={Proceedings of the IEEE/CVF International Conference on Computer Vision},
  pages={1216--1227},
  year={2025}
}

@article{alkhatib2025conformal,
  title={On Conformal Machine Unlearning},
  author={Alkhatib, Yahya and Tay, Wee Peng},
  journal={arXiv preprint arXiv:2508.03245},
  year={2025}
}

@article{li2025quantifying,
  title={Quantifying Uncertainty in Natural Language Explanations of Large Language Models for Question Answering},
  author={Li, Yangyi and Huai, Mengdi},
  journal={arXiv preprint arXiv:2509.15403},
  year={2025}
}

@inproceedings{
wei2024evaluating,
title={Evaluating Copyright Takedown Methods for Language Models},
author={Boyi Wei and Weijia Shi and Yangsibo Huang and Noah A. Smith and Chiyuan Zhang and Luke Zettlemoyer and Kai Li and Peter Henderson},
booktitle={The Thirty-eight Conference on Neural Information Processing Systems Datasets and Benchmarks Track},
year={2024},
url={https://openreview.net/forum?id=ar8aRMrmod}
}

\end{document}